\documentclass[sigconf,nonacm]{acmart}

\AtBeginDocument{%
  \providecommand\BibTeX{{%
    \normalfont B\kern-0.5em{\scshape i\kern-0.25em b}\kern-0.8em\TeX}}}

\newcommand{\tool}{AutoHistograms}
\newcommand{\company}{Google}

\setcopyright{acmlicensed}
\copyrightyear{2024}
\acmYear{2024}
\acmDOI{XXXXXXX.XXXXXXX}

\acmConference[CHI '24]{Make sure to enter the correct
  conference title from your rights confirmation emai}{May 11--16,
  2024}{Honolulu, HI}
\acmBooktitle{CHI '24: ACM conference on Human Factors in Computing Systems,
May 11--16, 2024, Honolulu, HI}
\acmISBN{978-1-4503-XXXX-X/18/06}

\usepackage{subcaption} 
\usepackage{enumitem}

\begin{document}

\title[Automatic Histograms]{Automatic Histograms: Leveraging Language Models for Text Dataset Exploration}

\settopmatter{authorsperrow=4, printfolios=true}

\author{Emily Reif}
\authornote{Both authors contributed equally to this research.}
\email{ereif@google.com}
\orcid{0000-0003-3572-6234}
\affiliation{%
  \institution{Google Research}
  \city{Seattle}
  \state{WA}
  \country{USA}
}

\author{Crystal Qian}
\orcid{0000-0001-7716-7245}
\authornotemark[1]
\email{cjqian@google.com}
\affiliation{%
  \institution{Google Research}
  \city{New York City}
  \state{NY}
  \country{USA}
}

\author{James Wexler}
\email{jwexler@google.com}
\affiliation{%
  \institution{Google Research}
  \city{Cambridge}
  \state{MA}
  \country{USA}
}

\author{Minsuk Kahng}
\orcid{0000-0002-0291-6026}
\email{kahng@google.com}
\affiliation{%
  \institution{Google Research}
  \city{Atlanta}
  \state{GA}
  \country{USA}
}


\begin{abstract}
Making sense of unstructured text datasets is perennially difficult, yet increasingly relevant with Large Language Models. Data workers often rely on dataset summaries, especially distributions of various derived features. Some features, like toxicity or topics, are relevant to many datasets, but many interesting features are domain specific: instruments and genres for a music dataset, or diseases and symptoms for a medical dataset. Accordingly, data workers often run custom analyses for each dataset, which is cumbersome and difficult. We present \tool{}, a visualization tool leveraging LLMs. \tool{} automatically identifies relevant features, visualizes them with histograms, and allows the user to interactively query the dataset for categories of entities and create new histograms. In a user study with 10 data workers (n=10), we observe that participants can quickly identify insights and explore the data using \tool, and conceptualize a broad range of applicable use cases. Together, this tool and user study contribute to the growing field of LLM-assisted sensemaking tools.

\end{abstract}


\begin{teaserfigure}
  \includegraphics[width=\textwidth]{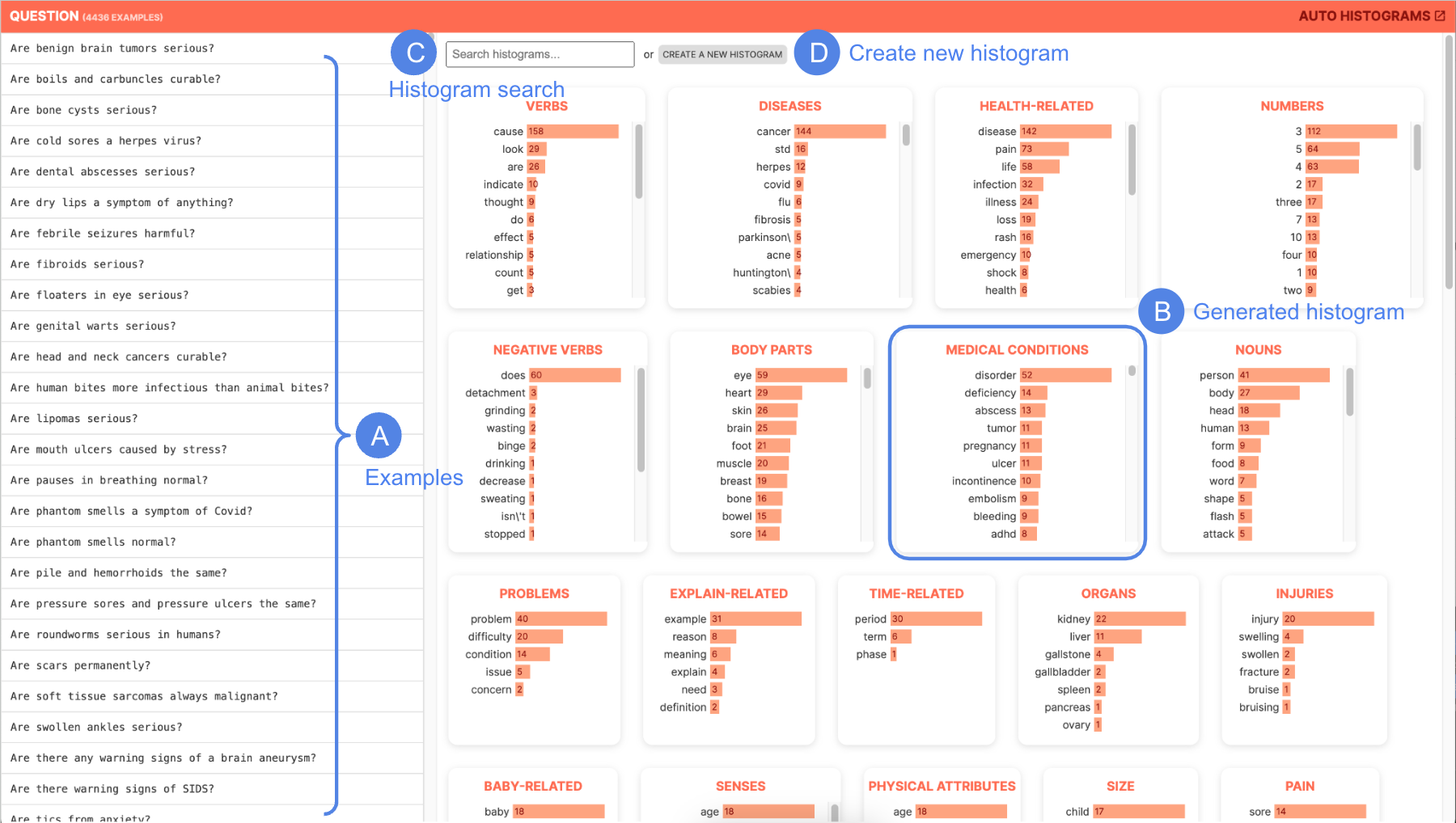}
  \caption{\textit{\tool{}: A tool for making sense of unstructured text datasets.} \textbf{(A)} Dataset examples are shown on the left hand side. \textbf{(B)} Dataset-specific distributions of entities in the dataset are generated in a pre-processing step, and visualized with histograms. \textbf{(C)} Histograms can be searched with exact string matching or semantic search. \textbf{(D)} New distributions can be generated in real time for in-the-loop dataset exploration.}
  \label{fig:teaser}
\end{teaserfigure}

\received{25 January 2023}

\maketitle

\section{Introduction}
\label{sec:intro}
Making sense of unstructured text datasets is an increasingly important, unsolved challenge. There are many high-stakes use cases where it is essential \cite{sambasivan}, especially with the rise of large language models (LLMs). These include curating the pre-training and finetuning datasets of LLMs, and creating evaluation benchmark datasets for areas such as safety, factuality, or other desired behaviors. Ideally, there would be quantitative methods to determine if a dataset is high quality. However, given that many of these LLM tasks are open-ended (e.g., creative writing, summarizing, or question answering), standard accuracy metrics can be inappropriate or insufficient \cite{Gehrmann2022RepairingTC}, as there is often no ground truth at all. To determine if a dataset is of sufficient quality, the data worker must first define what quality means in the context of the dataset. To do this, they must qualitatively understand the dataset itself.

As it is usually impossible to read every example in a dataset \cite{zhou2017}, many of the analyses for understanding unstructured text datasets center around calculating distributions and diversity \cite{Lara2022EvaluationOS} along specific derived features of the text. The field is converging on which features are applicable across many datasets (e.g., toxicity, topics, or protected groups) and formalizing them into frameworks \cite{gebru2021datasheets, pushkarna2022data, diaz2023sound}. There are also pipelines \cite{perspective, elazar2023whats} and visualization tools~\cite{kyd, ibm, datameasurementstool, lilac} to annotate and explore these broadly-applicable features. These tools go a long way toward supporting dataset understanding. However, they are not enough on their own; in practice, the analyses or features that data workers often care most about are dataset-specific. Alternative approaches to text dataset analysis include unsupervised techniques (e.g., topic modeling \cite{10.5555/944919.944937}, and dimensionality reduction \cite{10.5555/3454287.3455058}), and there are tools \cite{brath2023} that visualize these topics or other unsupervised clusters of examples
~\cite{kucher2015text, choo2013utopian, assogba2023large}. However, the features they produce can be hard to interpret.


To address these challenges, we present \tool{}, a visualization tool that automatically extracts semantically-meaningful attributes from raw unstructured text, and displays interactive visualizations of their distributions in the form of bar charts. \tool{} is open source\footnote{https://github.com/PAIR-code/auto-histograms}, and leverages LLMs' generative abilities and rich embedding spaces to cluster domain-specific attributes. Given a dataset, it automatically creates histograms of specific features relevant to that dataset. For example, if the dataset contains mentions to ``covid 19'', ``the flu'', and ``SARs'', \tool{} groups these terms and produces a histogram of examples with ``infectious diseases'' (see Section \ref{sec:preprocessing}). Users can also easily create new histograms in real time for in-the-loop hypothesis testing. For example, they can query a dataset with the natural language description of ``body parts'' without having to define a preset list of all possible body parts to run the analysis (see Section \ref{sec:creation}).

Note that \tool{} specifically addresses the issue of finding more dataset-specific distributions. We expect that it will be used in conjunction with more general annotation tools that annotate some of the features described above. 

We also present a user study with 10 data workers and data tool creators to evaluate \tool{}. Participants ramped up quickly and were able to perform actions defined within our key user journeys with minimal assistance. They were able to flag contextually relevant features of the dataset using the tool, and identified opportunities to apply \tool{} in a range of other use cases, including verifying safety, detecting outliers, debiasing example selection, and identifying mode collapse in synthetic data.

\section{User Challenges}

The following user challenges are based on previous informal conversations with data workers at \company{} where the authors make tools for evaluating training, benchmark, or synthetically generated data. Prior literature has already identified a range of common user needs around navigating a dataset \cite{Amar2005LowlevelCO}. We focus on the challenges of users who finetune and evaluate LLMs, and thus need to develop a qualitative understanding of unstructured text datasets.

\begin{itemize}
    \item \textbf{C1: Summarize the dataset with relevant distributions.} As mentioned in Section \ref{sec:intro}, one common practice is to look at distributions of derived text attributes. As annotating and analyzing dataset-agnostic attributes is already well-supported, we focus on categorical features that are \textit{specific to a given dataset}. For example, someone curating a dataset for a music recommendation system might care about the distribution of genres, instruments, or artists, but someone making a more targeted responsible AI benchmark dataset might care about the distribution of specific religions, genders, or races. There are a few specific steps in this process:
        \begin{enumerate}[label=\alph*)]
        \item \textbf{Determining relevant features.} In a novel dataset, it is not always immediately clear what features will provide interesting insights in the data.
        \item \textbf{Annotating identified features.} It can be difficult to annotate data with these features, especially at scale.
        \item \textbf{Displaying the feature distributions.} After each example is annotated with the feature, it is necessary to have some form of visualization or summary of the feature across the dataset. 
        \end{enumerate}
\item \textbf{C2: Find pathological distributions.} While this shares many of the low-level implementation challenges as \textbf{C1}, finding imbalances in the derived features is often described as a separate high-level goal.

\item \textbf{C3: Find surprising slices of data.} Complementary to the summary, users also need to find groups of examples that wouldn't necessarily be captured in the main summaries or by a quick scan.  For example, a medical dataset might have a group of examples suggesting fringe medical advice. There might be so few examples that these would not be found by either a quick overview of the dataset, or in the main summaries.

\item \textbf{C4: Onboard quickly.} While not directly related to dataset understanding, an essential need that is often overlooked is being able to actually use these tools without too much startup cost. Ideally, these tools would be automatically integrated into standard workflows.

\end{itemize}

\begin{figure*}
  \includegraphics[width=\textwidth]{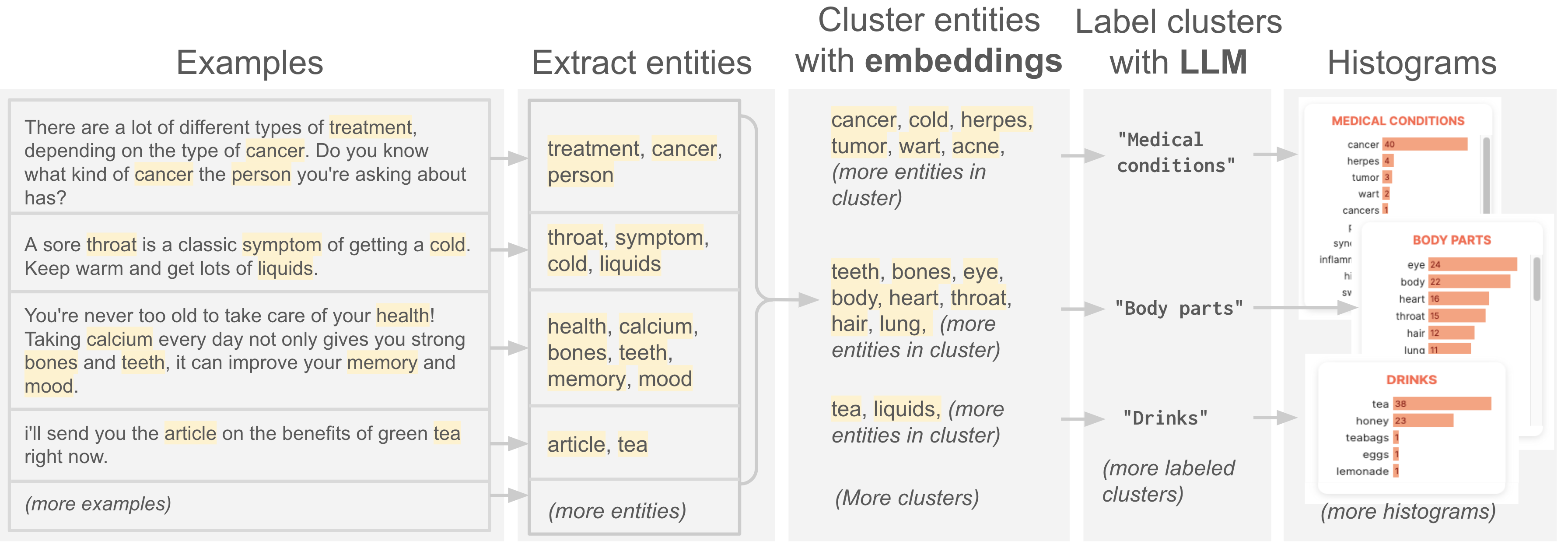}
  \caption{The processing steps for automatically creating histograms from a dataset}
  \label{fig:steps}
\end{figure*}

\section{Design Goals}
To address these unmet needs, we designed our tool with the following goals in mind.

\begin{itemize}
    \item \textbf{G1: Automatically select salient attributes.} To address \textbf{C1} and \textbf{C2}, \tool{} must determine what features are relevant to a dataset. To address \textbf{C4}, this must be mostly automatic; the user should not have to pre-define every feature themselves.
    \item \textbf{G2: Let users quickly iterate on these attributes.} However, this automatic feature selection will not always be perfect. Also in support of \textbf{C1}, users should be able to add to the automatically-generated histograms by creating new ones in real time. This free form exploration also supports \textbf{C2}.

    \item \textbf{G3: Visualize attributes distributions} To support \textbf{C1-3}, the tool should display the feature distributions in an easily digestible format. For \textbf{C3}, specifically, it should let the user interactively dig into the specific examples that belong to a bucket of a given histogram.

\end{itemize}

\section{Visualization design and implementation}

\tool{} has two components:
\begin{enumerate}
    \item A pre-processing pipeline to create the histograms
    \item A visualization tool for viewing the generated histograms, and creating more interactively
\end{enumerate}

\subsection{Preprocessing pipeline: creating the histograms}
\label{sec:preprocessing}

\subsubsection{Extract entities}
The first step of the pipeline is to collect all entities across the dataset. We use NLTK \cite{bird2009natural} to select the nouns and numbers in the dataset. For performance reasons, we keep the most frequent k=2000 entities.
\subsubsection{Cluster entities with embeddings} We then find meaningful groups of entities to create the histograms. To do this, we calculate the embedding of each entity using the PaLM API \footnote{https://ai.google/discover/palm2/ , text-gecko model}, then cluster the entities in the embedding space. A given entity (e.g., "email") can be in multiple histograms (e.g., "communications", "computer-related"), so we run hierarchical clustering\footnote{https://docs.scipy.org/doc/scipy/reference/cluster.hierarchy.html} using a range of values for the cluster similarity cutoff to get many clusters. 

\subsubsection{Label clusters with LLMs}
We use PaLM API\footnote{https://ai.google/discover/palm2/} to label the groups of entities using a few shot prompt. We also include a `no label' category for low quality clusters. The resulting labeled list of entities can then be converted to a histogram by simply counting the number of examples in the dataset that contain the entity.

\subsection{Interactive exploration}
\begin{figure*}
    \includegraphics[width=\textwidth]{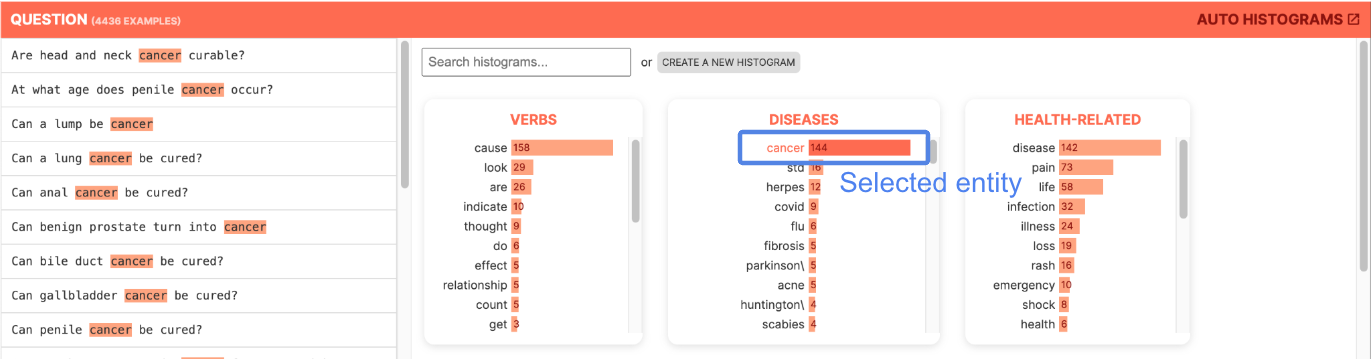}
    \caption{When the histogram bucket for an entity is selected, the data table is filtered to only examples that contain that entity.}
    \label{fig:selected}
\end{figure*}

The UI (Figure \ref{fig:teaser}) allows the user to interactively explore the histograms and create new ones. The left side (Figure \ref{fig:teaser}(A)) is a scrollable list of examples. The right side contains the automatically generated histograms (Figure  \ref{fig:teaser}(B)). When the user selects an entity in a histogram  (Figure \ref{fig:selected}), the entity is highlighted, and the data table is filtered to only show examples that contain that entity.
The user can also search for histograms by exact or semantic similarity (e.g., searching ``diseases'' also surfaces ``illnesses''.) The UI is implemented using TypeScript and the LIT framework.\footnote{https://lit.dev}

\begin{figure}
     \includegraphics[width=\linewidth]{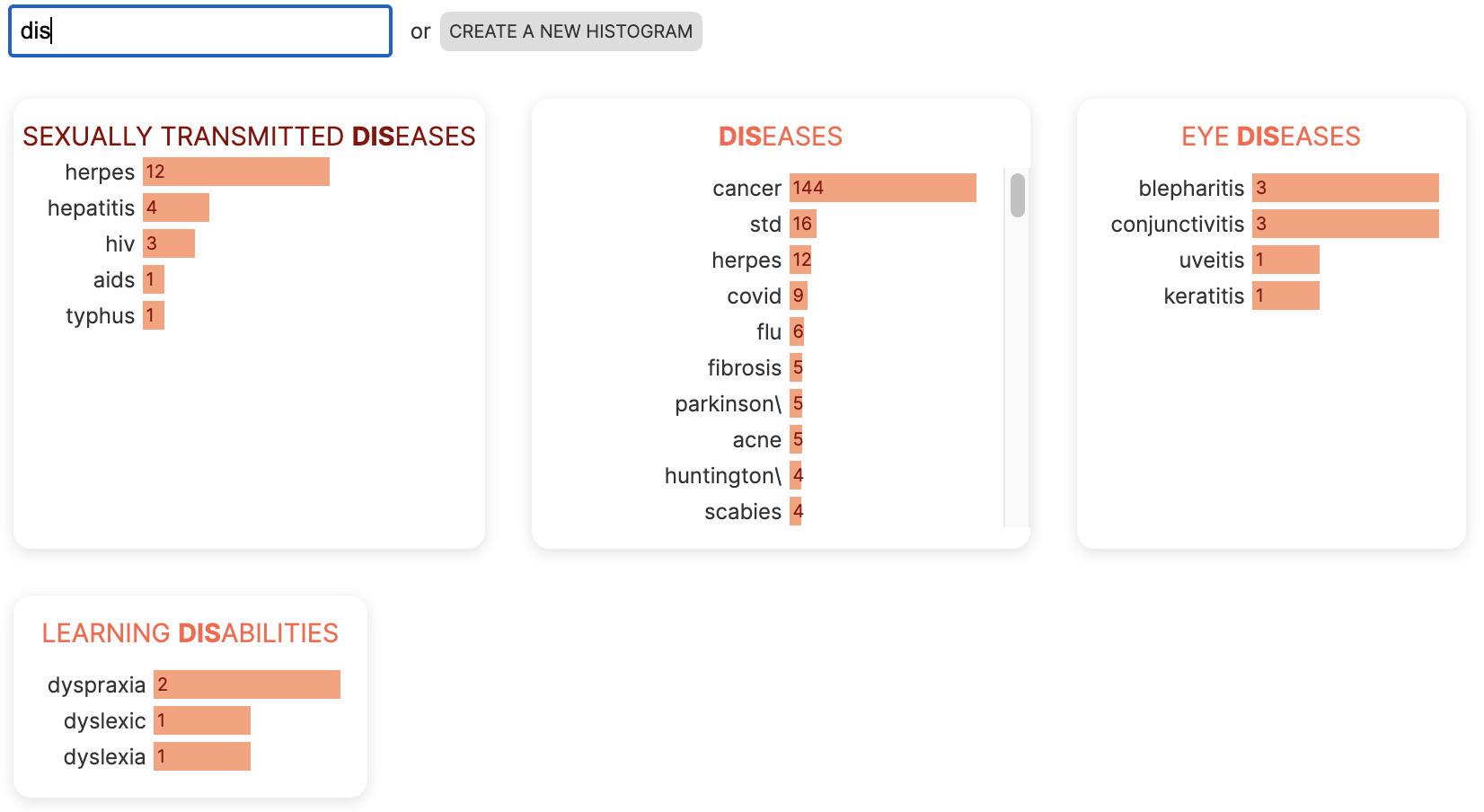}
    \caption{A search interface that supports exact or semantic search.}
\end{figure}

\begin{figure*}
    \includegraphics[width=\linewidth]{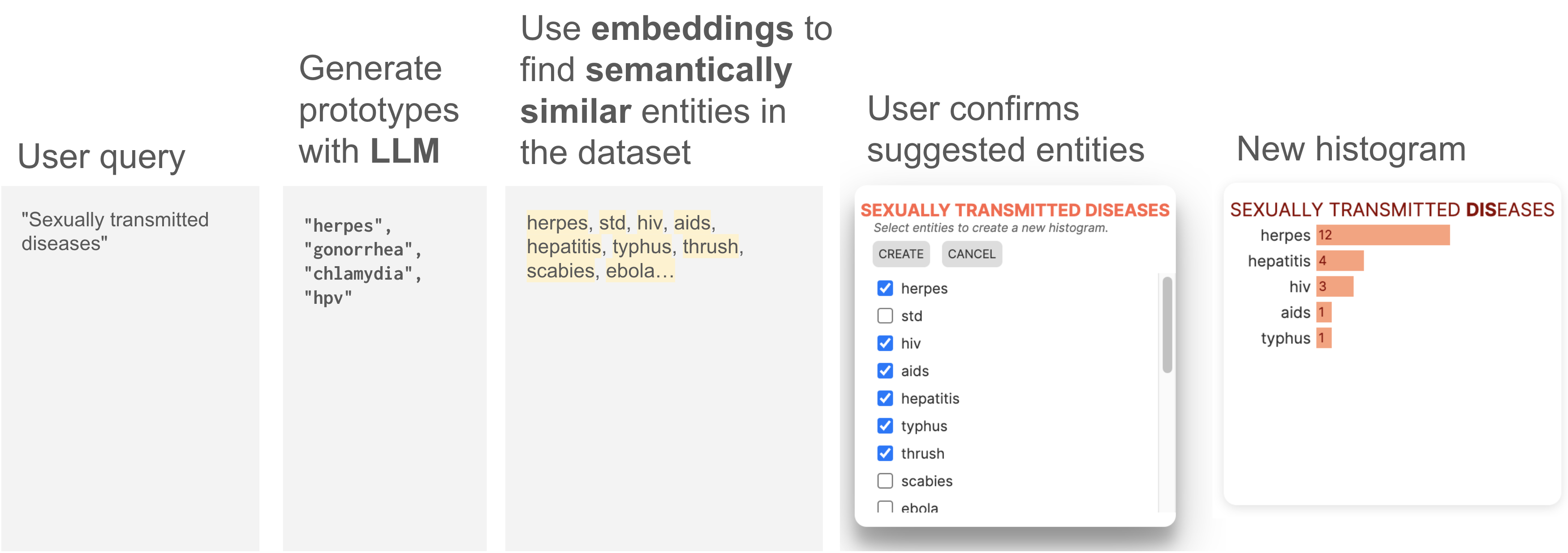}
    \caption{The steps for creating a new histogram on-the-fly. The user types in a category of entities they wish to find in the dataset (e.g., "sexually transmitted diseases"). Using LLMs and embeddings, we surface entities in this category, which the user can confirm to create a new histogram.}
\end{figure*}
\label{fig:search_and_create}

\subsection{Real time creation of new histograms}
\label{sec:creation}
  
If the user would like to explore a feature that was not automatically generated as part of the pipeline, they can create a new histogram in real time with a human-in-the-loop process (see Figure \ref{fig:search_and_create}) leveraging LLMs and embeddings:
\begin{enumerate}
    \item User types query (e.g., "sexually transmitted diseases").
    \item The LLM is queried to "give me examples of <new feature name, e.g. sexually transmitted diseases>". It returns some entities, which may or may not actually be in the dataset.
    \item Given these LLM-generated entities, we suggest semantically similar entities that are in the dataset by calculating the centroid of the LLM-generated entities in the embedding space. We then surface the entities in the dataset that have high embedding cosine similarity to this centroid.
    \item These semantically similar entities are then presented to the user, who selects those they wish to include in the histogram.
    \item Finally, we create the histogram of examples with each entity, and integrate it back into the UI.

\end{enumerate}

\section{Observational Study}
The structure of each individual, 30-minute, virtual user study is as follows:

\begin{itemize}
    \item \textbf{Introduction} (5 min): Participant describes their background and use cases
    \item \textbf{Demo} (5 min): Moderator briefly demonstrates \tool{} on a sample dataset of musical terms, including the search feature and histogram creation
    \item \textbf{Free-form exploration} (20 min): Participant follows a link to the tool and explores a dataset of sample chatbot responses to medical queries, sharing their screens with the moderator and thinking aloud. During this free-form exploration, participants took different angles of their choice, such as deciding whether to evaluate this dataset for safety concerns or cleaning the dataset for any apparent outliers to train a large language model. 
\end{itemize}

\subsection{Participants}

\begin{table*}
\resizebox{.8\linewidth}{!}{%
\begin{tabular}{ c l  }
 \toprule
 \textbf{Participant} & \textbf{Experience} \\
 \midrule
 P1 & Builds data pipelines for training conversational LLMs \\ 
 P2 & Generates synthetic data for adversarial testing of LLMs \\ 
 P3 & Builds tools for curating and annotating text-based datasets \\ 
 P4 & Develops text-to-image models using multimodal datasets \\ 
 P5 & Builds tools for curating and interpreting text-based datasets \\ 
 P6 & Conducts mix-methods research on data annotation agreement \\ 
 P7 & Uses LLMs to automatically rate benchmarking datasets for debugging model behaviors \\ 
 P8 & Conducts qualitative research on data annotation subjectivity \\ 
 P9 & Develops tools for improved data labeling and understanding \\ 
 P10 & Builds text-based datasets from webpage sources \\ 
 \bottomrule
\end{tabular}
}
  \caption{A summary of participants and relevant experience.}

\label{tab:participants}
\end{table*}

We recruited 10 industry professionals at a large technology company (N=10) who have experience interacting with text-based datasets. Table \ref{tab:participants} summaries their relevant experience. These participants included tool developers, model developers, and research scientists. Most of the data they interact with are for the purpose of training and evaluating large language models. As part of the background interview, we asked participants what visualization tools they used to understand their data. For the most part, they are not currently using visualization techniques, but they want to (emphasizing \textbf{C1c}).
Although they occasionally use tools like Jupyter or Colab, data exploration is usually performed by visually scanning examples in a .csv or spreadsheet. 
Most participants would like to use and explore visualization tools further, but do not due to the difficulty of easily creating them.

    \begin{quote}
        \textit{``I probably would [look at visualizations] if I took the time to build that. I tried to create a pie chart to show the distribution, but I just left it as a table. I was too lazy.''}\hspace{1em plus 1fill}---P2
    \end{quote}
    
\subsection{Observations}
We conducted a thematic analysis \cite{braun} to analyze and code behaviors and commentary from the user sessions. Across the 10 user studies, we found evidence that the \tool{} addressed all four targeted user challenges. 

\textbf{C1, Identifying subject matter}: Participants were not told the subject matter of this dataset. However, all participants correctly quickly identified the subject as \textit{chatbot responses to medical queries}. They accomplished this by referencing the histogram panel and then validating in the data panel, rather than manually scrolling through the list of examples: 9/10 users initially focused on the histograms rather than the list of examples. Participants appeared to heavily rely on and interact with the histograms panel to synthesize their insights; they would refer to the list of examples primarily only to validate hypotheses and insights generated from looking at the histograms panel.

\textbf{C2, Finding pathological distributions}: Participants spent more time studying histograms with higher entropy, demonstrated by hovering, scrolling, and clicking on these long-tailed histograms. Six participants commented on the \textit{diseases} histogram in Figure \ref{fig:teaser}, which had 144 instances of the token "cancer," 130 more instances than than the second-most frequent token.

\begin{quote}
    \textit{```I’m  interested in surprises- for example, long tails. There’s lots of cancer but not other medical conditions.''}\hspace{1em plus 1fill}---P3
\end{quote}

\begin{quote}
    \textit{``Say I wanted to create a dataset that's balanced across diseases.. this tells me that it's [focused on] cancer. These titles [of histograms] tell me everything I need to know about the dataset.''}\hspace{1em plus 1fill}---P3
\end{quote}

\textbf{C3, Identifying unexpected slices of data}:

The histograms are sorted by the total count of number of occurrences. This can cause seemingly-arbitrary concepts to appear at the top of the interface; for example, in Figure \ref{fig:teaser}, histograms about \textit{verbs} and \textit{numbers} appear next to \textit{diseases} and \textit{health}. The participants who were on the tool-building side appeared to be skeptical about the relevance of histograms for more ambiguous terms such as \textit{things} and \textit{ways}. However, the participants who performed more data analysis in their work reported liking that the histograms did not appear to be completely relevant. Adding a feature to specify sort order may help users to parse their data in ways to suit their needs. A common theme is that users wanted to be surprised by outliers; entropy was suggested three times as a sorting mechanism.

\begin{quote}
    \textit{``It’s neat that it’s surfacing relevant tokens.. But not all histograms are useful.''}\hspace{1em plus 1fill}---P5, a tool builder
\end{quote}

\begin{quote}
    \textit{``I like that there are seemingly less relevant suggestions of histograms (e.g. question words) because there can be surprising things. It was helpful, but I don’t [typically] think that way.''}\hspace{1em plus 1fill}---P6, a data scientist
\end{quote}

\begin{quote}
    \textit{``It's difficult to see where to look. You might look at 10 different directions and still nothing comes out until the 11th direction. We have potentially a hunch on what would be interesting to look at.. But we [are looking to be] surprised by what we see.''}\hspace{1em plus 1fill}---P7
\end{quote}

Participants were able to quickly identify and select interesting slices of data. Participants, not only those who worked on AI safety, wanted to ensure that the chatbot was not giving unsafe advice. Four participants typed the word ``advice'' into the search bar or created a new histogram of ``advice''-related terms to explore examples with this term. Using this workflow, participants were able to quickly flag potentially problematic examples of chatbot responses, such as "I'm not a doctor, but lemon and tea usually work for me."

\begin{quote}
    \textit{``[Thinking about safety] is required.. especially in generative AI, there's strict review to make sure that your generated information is actually safe.''}\hspace{1em plus 1fill}---P4
\end{quote}

\begin{quote}
    \textit{``This is just something borderline unsafe.. you're not supposed to give medical advice. For the bot to say that `I would not want to give you medical advice, but... That is a safety violation.''}\hspace{1em plus 1fill}---P8
\end{quote}

\begin{quote}
    \textit{``I’m hoping that [I’m not seeing] chatbot interactions related to health concerns because that would be against [company] policies.''}\hspace{1em plus 1fill}---P5
\end{quote}

\textbf{C4: Onboarding and hypothesis-testing quickly}: All participants were able to independently accomplish the above tasks with largely no intervention from the moderator. Participants were eager to actively interact with the tool; participants ubiquitously clicked on histogram bars and expected the relevant data panels to appear on the left data panel. Rather than passively view the histogram panels, they actively scrolled, clicked on bars, typed in search queries, and interacted with the UI to address their dataset hypotheses.

\begin{quote}
    \textit{``I immediately want to click this [bar] and see how many times it [appears]...''}\hspace{1em plus 1fill}---P7
\end{quote}

\subsection{Use cases}
Participants identified the following use cases as opportunities to integrate this tool into their existing workflows:

\textbf{Classification/tagging}: Participants voiced that \tool{} could help them to understand the contents of a large dataset quickly and with less bias than their current method of manually reading a few select examples.

\begin{quote}
    \textit{``If you had no prior information about the data... instead of reading all of the individual examples, you can read these [histogram lists]. Without this, I would have shuffled this data and then read it. I would have read a couple hundred before my eyes started bleeding.''}\hspace{1em plus 1fill}---P9
\end{quote}

\textbf{Misclassification/identifying outliers}: Using their current methods, data practitioners need to formulate hypotheses about existing outliers/bad data before finding them. By grouping tokens into buckets, \tool{} allowed users to quickly identify high-spread distributions such as cancer vs. other ailments.

\begin{quote}
    \textit{```We have these datasets that are supposedly of good quality. If you eyeball random examples, you can see that it's wrong, but you don't know how widespread that is in your dataset.''}\hspace{1em plus 1fill}---P7
\end{quote}

\textbf{Safety}: In terms of specific use cases, almost all participants identified \textit{safety} as an area where \tool{} could have a positive impact. Particularly as generative models have become more pervasive, our participants stressed the importance of making sure that models are trained on safe data. \tool{} could help to identify correlations that appear in harmful queries (P2) and identify subsets of the finetuning data to rebalance such that toxicity scores are below a compliant threshold (P6). \tool{} could help to label generated content that violate safety standards (such as by giving medical or legal advice), and identify sensitive or adversarial topics (e.g. religion, politics) (P7).

\textbf{Fairness}: Fairness was another common use case: data can be rebalanced to ensure better representation amongst subgroups (P4) and models can be fine-tuned with evenly-distributed synthetic data if biases are discovered (P8). 

\textbf{Synthetic data}: \tool{} could be used to identify mode collapse in synthetically generated data (P7).    

\subsection{Limitations and challenges}
Finally, we discuss the most common user feedback on the tool's limitations and challenges.

\textbf{1. Participants also want to explore numerical metadata}:
As discussed in the introduction, there are general (non-dataset-specific) features that are useful for analysis. Three participants (P1, P4, P7) said they might look at numerical features such as text length, number of examples, token counts, and summary statistics. They suggested that histograms of relevant statistics rather than text features may be more useful for their use cases.

\textbf{2: Demand for intersectional slicing}:
Balancing skews and uneven distributions of data appeared to be a key use case for data practitioners. Many participants wanted more fine-grained intersectional exploration in the tool, and asked for advanced searching (such as ``AND'' and ``OR'') clauses to support this need:

\begin{quote}
   \textit{``[We’re interested in] identity terms and formality of language. What kinds of topics come up [for different subgroups of annotators]? What is or is not being represented?''}\hspace{1em plus 1fill}---P6, on social-cultural context for annotator agreement
\end{quote}
\begin{quote}
    \textit{``The way that speaking is gendered is very subtle.. so [I’d want to categorize by] by word type, verbs.''}\hspace{1em plus 1fill}---P9, on representation in conversational datasets
\end{quote}

Participants also listed integration, speed, and reliability as key factors that would help them adopt \tool{}.

\section{Conclusion}
We present \tool{}, a tool that leverages LLMs and embeddings to create an interactive interface of automatically-generated histograms for data practitioners to analyze unstructured datasets. Through an observational study with 10 data workers, we validate that the tool can address targeted user needs such as summarizing datasets, identifying outliers and interesting slices of data, and testing hypotheses rapidly and interactively. Participants quickly identified the correct dataset topic, noticed a potentially-concerning asymmetrical data distribution, and found safety violations within the dataset. Finally, we summarize potential use cases and limitations of \tool{} described by these participants. Together, these findings suggest that advancements in LLMs can enable the development of sensemaking tools to better serve data workers.

\section{Acknowledgements}
The authors wish to thank our colleagues at Google's People + AI Research Team for helpful feedback and discussions, especially Lucas Dixon, Andy Coenen, Martin Wattenberg and Fernanda Vi\'{e}gas.

\bibliographystyle{ACM-Reference-Format}
\bibliography{sample-base}

\appendix

\end{document}